\documentclass[letterpaper, 10 pt, conference]{ieeeconf}
\usepackage{algorithm}

\IEEEoverridecommandlockouts                          
\usepackage{amsmath} 
\usepackage{amssymb}  

\usepackage{algpseudocode}
\usepackage{graphicx}
\usepackage{physics,bm,bbm}

\usepackage{multirow}
\usepackage{cuted}
\usepackage{cite}

\usepackage{booktabs}
\usepackage{colortbl}
\usepackage{multicol}
\usepackage{xcolor}
\usepackage{makecell}
\usepackage{subcaption}
\usepackage{enumitem}
\usepackage{float}
\usepackage{placeins}

\newcommand{\hide}[1]{}

\newcommand\blfootnote[1]{%
  \begingroup
  \renewcommand\thefootnote{}\footnote{#1}%
  \addtocounter{footnote}{-1}%
  \endgroup
}

\definecolor{es-blue}{rgb}{0,0.4,0.8}

\definecolor{lightgray}{gray}{0.9}

\definecolor{cvprblue}{rgb}{0.21,0.49,0.74}
\usepackage[pagebackref,breaklinks,colorlinks=true,citecolor=cvprblue, urlcolor=blue]{hyperref}

\begin{document}

\title{\LARGE \bf
ACE-F: A Cross Embodiment Foldable System with \\ Force Feedback for Dexterous Teleoperation
}


\author{
    Rui Yan\textsuperscript{*} \quad
    Jiajian Fu\textsuperscript{*} \quad
    Shiqi Yang\textsuperscript{*} \quad
    Lars Paulsen\textsuperscript{*} \quad
    Xuxin Cheng \quad
    Xiaolong Wang \\
    UC San Diego
    \thanks{* Equal contributions.}
}

\twocolumn[{%
\renewcommand\twocolumn[1][]{#1}%
\maketitle

\begin{center}
    \vspace{-0.1in}
    \centering
    \captionsetup{type=figure}
    \includegraphics[width=1\linewidth]{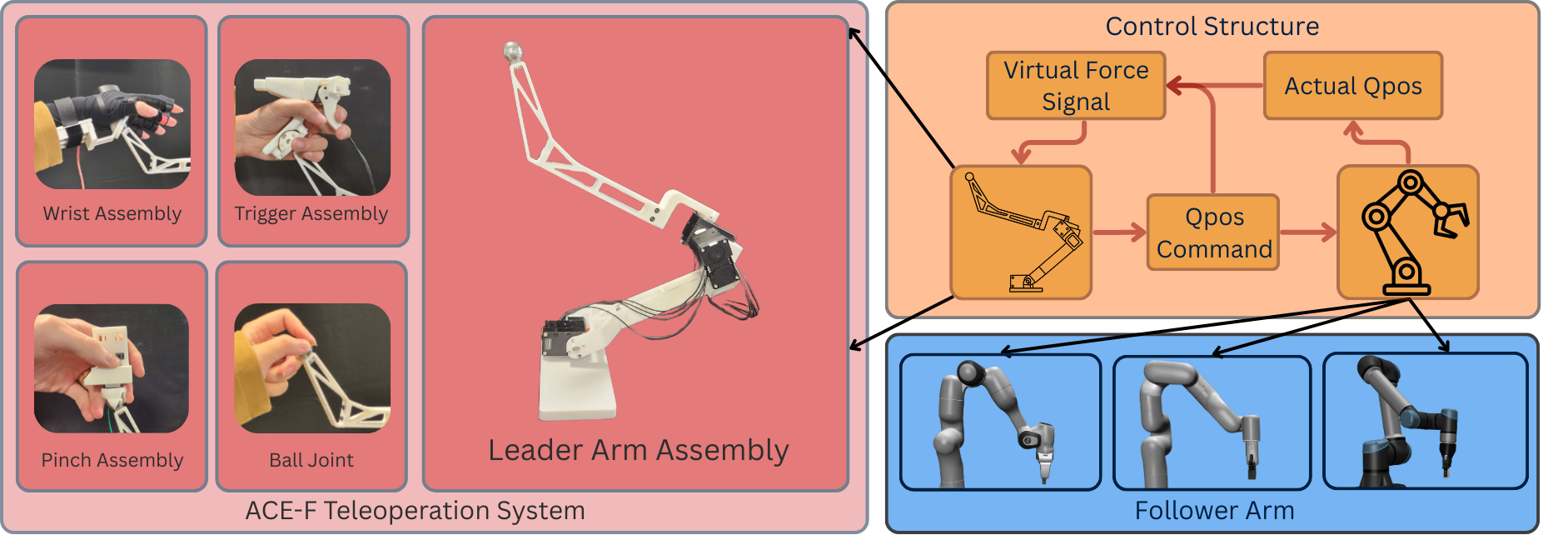}
    \caption{ACE-F enables teleoperation users to complete contact-rich tasks on a wide range of simulated and real-world robots using an innovative virtual force feedback model and highly simplified mechanical design.}
    \label{fig:intro_schem}
    \vspace{-0.1in}
\end{center}
}]

{\blfootnote{{\authorrefmark{1} Equal contribution. }}}

\thispagestyle{empty}
\pagestyle{empty}

\begin{abstract}
Teleoperation systems are essential for efficiently collecting diverse and high-quality robot demonstration data, especially for complex, contact-rich tasks. However, current teleoperation platforms typically lack integrated force feedback, cross-embodiment generalization, and portable, user-friendly designs, limiting their practical deployment. To address these limitations, we introduce ACE-F, a cross embodiment foldable teleoperation system with integrated force feedback. Our approach leverages inverse kinematics (IK) combined with a carefully designed human-robot interface (HRI), enabling users to capture precise and high-quality demonstrations effortlessly. We further propose a generalized soft-controller pipeline integrating PD control and inverse dynamics to ensure robot safety and precise motion control across diverse robotic embodiments. Critically, to achieve cross-embodiment generalization of force feedback without additional sensors, we innovatively interpret end-effector positional deviations as virtual force signals, which enhance data collection and enable applications in imitation learning. Extensive teleoperation experiments confirm that ACE-F significantly simplifies the control of various robot embodiments, making dexterous manipulation tasks as intuitive as operating a computer mouse. The system is open-sourced at: \url{https://acefoldable.github.io/}
\end{abstract}
\section{Introduction}
\label{sec:intro}
Teleoperation systems have shown great potential for collecting high-quality, diverse demonstration data for complex, contact-rich robotic tasks. However, existing platforms have three main limitations: (1) lack of integrated force feedback—either providing no haptic cues or relying on expensive, hard-to-integrate force/torque (FT) sensors \cite{imdieke2025spark,ma2025fewshot}; (2) poor cross‐embodiment generalization—joint‐copying schemes must be redesigned for each new robot morphology \cite{lv2022gulim,hejrati2025teleop}; and (3) bulky, non-portable hardware that hinders rapid deployment in real-world scenarios \cite{zhao2023wearable,hejrati2025teleop}. High component costs further complicate these teleoperation systems, which poses an additional challenge. To address these problems, we propose \textbf{ACE-F}, a cross-embodiment foldable teleoperation system with integrated force feedback. First, ACE-F infers real‐time 3-DoF external forces by monitoring end-effector (EE) trajectory deviations, no additional sensors required, and applies active gravity and friction compensation on the leader and follower arms to deliver smooth, intuitive haptic cues \cite{khalifa2024sensorless,ma2025fewshot}. Second, we combine IK–based leader-arm control with glove-based hand tracking to build a universal EE retargeting algorithm that adapts to diverse robot platforms; a magnetic quick-swap interface further enables future integration of tactile gloves \cite{yang2024ace,leonardis2024glove, Zhang2025doglove}. Finally, our soft-controller pipeline fuses proportional-derivative (PD) control with custom inverse dynamics (ID), ensuring stability, responsiveness, and safety across embodiments, and allowing rapid deployment via minor tuning of URDF parameters \cite{yang2024ace}. ACE-F achieves high-precision demonstrations and maintains affordability and portability. It demonstrates two clear advantages in experiments: (1) users can rapidly adapt and accurately perform cross-platform teleoperation tasks under varying precision and workspace requirements; and (2) at a relatively low cost, ACE-F significantly outperforms systems without force feedback in complex contact-rich tasks.

\begin{figure*}[!t]
  \centering
  \includegraphics[width=\textwidth]{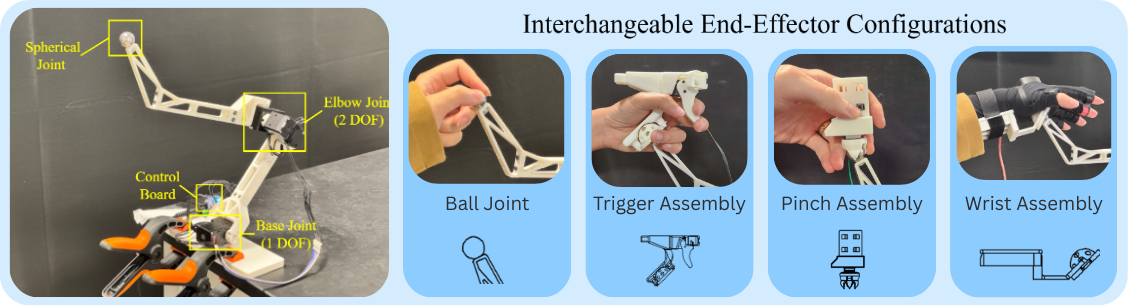}
  \caption{
    Overview of the ACE-F system. \textbf{Left}: Annotated view of the ACE-F arm showing the base joint (1 DoF), perpendicular elbow joints (2 DoF), and the spherical joint for interchangeable end-effectors. \textbf{Right}: Several representative end-effector configurations are enabled by the spherical joint.
  }
  \label{fig:acef-hardware}
\end{figure*}

\section{Method}
ACE-F makes innovations in four primary areas: hardware design, end-effector control, an augmented IK solver, and force feedback. The 3-DoF structure minimizes complexity, which simplifies control tasks while providing a wide range of movement, the IK solver prevents the system from entering singularities during long-horizon tasks, the end-effector avoids drift by connecting to the base while orientations are recorded with an inertial measurement unit (IMU), and virtual forces are calculated to provide users with an additional sense for contact-rich tasks. \FloatBarrier

\subsection{Hardware Design.} 
The ACE-F system, illustrated in Fig.~\ref{fig:acef-hardware}, is a robotic manipulator designed for precise 3-DoF force feedback. It features three independent joints: a base joint for foundational rotation and two perpendicular elbow joints for compact and robust force rendering. The manipulator uses DYNAMIXEL XM430-W350-R motors with U2D2 controllers, ensuring precise and responsive joint control. At its endpoint, an exposed ball-joint supports interchangeable end-effectors via a 3D-printed ball-and-socket design. As an extra precaution, an elastic safety lock mechanism prevents unintended detachment during operation. We validated three end-effector categories: A bare configuration which serves as a general-purpose setup for force feedback, two gripper configurations enabling simple gripper actuation, and a glove configuration tailored for humanoid platforms equipped with dexterous hands. This modular design significantly enhances the system's flexibility and adaptability for diverse real-world applications.

\subsection{Augmented Inverse Kinematics Solver.}
Unlike conventional IK solvers that purely minimize end-effector position and orientation errors, we propose an augmented IK approach tailored to the unique teleoperation challenges of over-actuated arms. Our solver introduces two additional “tasks” to improve robustness and avoid singularities. First, we compute the projection angle of the end-effector onto the base plane and match it to the first joint angle of the robot. This ensures a natural alignment between the direction the operator wants to face and the robot’s base rotation. Second, to prevent the extra joint in 7-DoF arms from bending outward when the end effector approaches the base, an action that can lead to kinematic singularities, we introduce a soft constraint on the fourth joint. We assign it a higher target value in the vertical ($z$) axis, encouraging a posture that avoids such configurations. These task-level modifications enhance the solver’s reliability and stability, allowing intuitive and continuous teleoperation even near the robot’s kinematic limits.

\begin{figure}[b]
  \centering
  \includegraphics[width=\columnwidth]{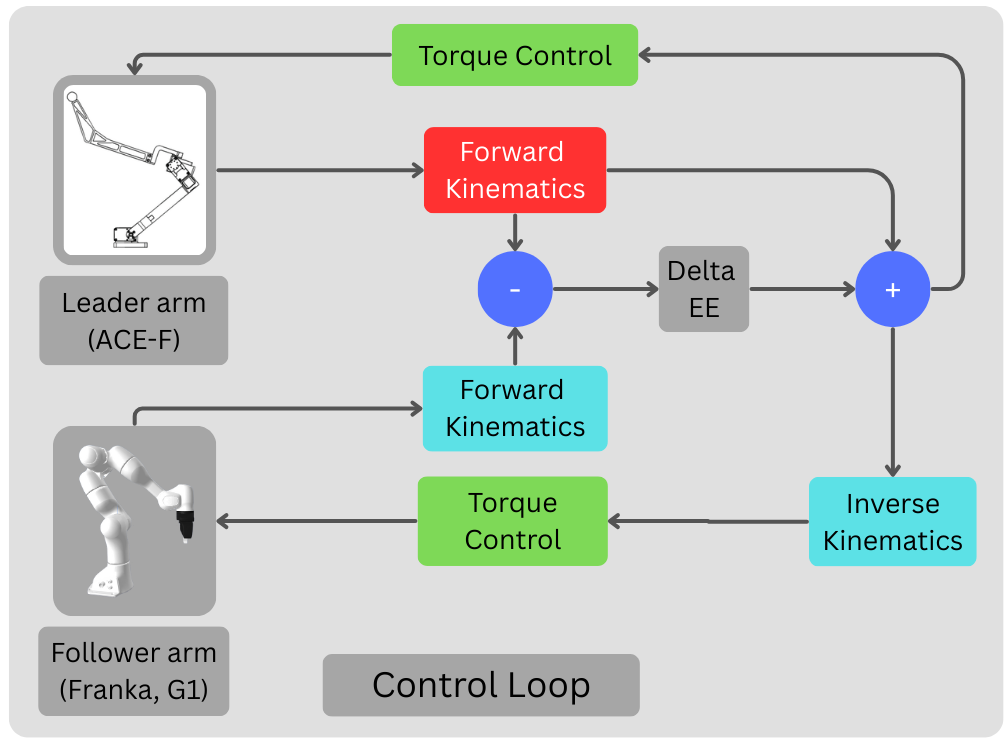}
  \caption{
    ACE-F's scaled virtual force-feedback control system.
  }
  \label{fig:control_struct}
\end{figure}

\subsection{End-Effector Control and Feedback.} 
One core challenge in teleoperation is achieving full 6-DoF control (position and orientation) of the robot end-effector using compact and low-DoF input devices. A 3-DoF arm alone cannot simultaneously define both the position and orientation of the end-effector in space. To address this limitation, ACE-F decouples position and orientation control: the foldable 3-DoF leader arm is used to determine the Cartesian position of the end effector, while an IMU captures the wrist orientation in real time. This can be expanded further to include finger orientations by integrating a glove-based tracking module. Combining these two streams, ACE-F reconstructs a complete 6-DoF in-hand pose of the operator, which can then be retargeted to the robot.

This hybrid control design preserves portability and affordability and enables platform-agnostic retargeting. The system uses IK to convert the desired position from the leader arm and the orientation from the glove into robot-specific end-effector commands. As a result, users can define both position and rotation of the end-effector naturally within the workspace, enabling seamless teleoperation across robots with different kinematic structures.

Additionally, ACE-F incorporates virtual force feedback by interpreting trajectory deviations between the commanded and actual end-effector positions as 3-DoF force signals, as shown in Fig. \ref{fig:control_struct}. These inferred forces are rendered on the leader arm via active torque control, alongside gravity and friction compensation. This provides intuitive haptic cues without requiring external sensors, by making contact events such as collisions, object slippage, or resistance perceptible to the operator.

ACE-F preserves the flexibility of the original ACE system, such as the ability to adapt to different workspace scales through simple geometric transformations and offers key advantages in force feedback implementation. By reducing the leader arm to 3-DoF, ACE-F simplifies mechanical design and real-time torque control. This reduction makes it significantly easier to implement reliable, low-latency force feedback, because the system only needs to estimate and render 3D translational forces, instead of full 6D wrenches. Together, these properties enable intuitive and high-fidelity teleoperation across diverse robotic platforms while maintaining portability, low cost, and modular expandability.

\subsection{Force Feedback Calculation.} Instead of relying on conventional Jacobian-based wrench-to-torque mapping for force feedback calculation, we compute the deviation between the follower arm end-effector’s target and actual positions:

\[
\Delta \mathbf{ee} = \mathbf{target} - \mathbf{current}
\]

This deviation, $\Delta \mathbf{ee}$, serves as the core indicator of the feedback force magnitude. Traditional wrench-to-torque mappings can be highly sensitive to transient forces, where short-duration impulses can result in substantial torque spikes that destabilize the leader arm. To circumvent this issue, we introduce a virtual target pose for the leader arm. By scaling $\Delta \mathbf{ee}$ and applying it to the leader arm’s current end-effector pose, the leader arm naturally tries to align with the follower’s pose via force feedback, as though the two end-effectors are connected by a virtual spring. Since the operator’s hand firmly grips the end-effector, this alignment manifests as tangible forces rather than significant positional displacements, thereby avoiding large oscillations. 

To further prevent destabilizing effects at high speed, where $\Delta \mathbf{ee}$ may inflate due to dynamic motion rather than contact, we regulate the feedback magnitude by the follower’s Cartesian velocity:

\[
\text{Force Feedback Factor} = \sqrt{\frac{\alpha \cdot \|\Delta \mathbf{ee}\|^2}{1 + \|\mathbf{v}_\text{cartesian}\|^2}}
\]

This ensures that the feedback force remains negligible during smooth motions, only becoming significant during actual contact interactions. Additionally, we apply this feedback factor not only to generate the virtual target pose but also to adaptively modulate the leader arm’s impedance gains ($K_p$ and $K_d$), implementing stable and intuitive haptic feedback across tasks.

\begin{algorithm}[ht]
\caption{Force Feedback–Enhanced Teleoperation Loop}
\label{alg:force_feedback}
\begin{algorithmic}[1]
\Repeat
    \State \textbf{Leader Arm:} Compute target $\mathbf{ee}_\text{target}$ based on operator’s 3-DoF arm pose and glove orientation
    \State \textbf{Leader Arm:} Solve IK for $\mathbf{q}_\text{target}$ of the follower arm
    \State \textbf{Leader Arm:} Send $\mathbf{q}_\text{target}$ to follower arm
    \State \textbf{Follower Arm:} Send current joint positions $\mathbf{q}_\text{current}$ to leader
    \State \textbf{Follower Arm:} Receive $\mathbf{q}_\text{target}$ and start moving toward it
    \State \textbf{Leader Arm:} Compute current follower $\mathbf{ee}_\text{current}$ using forward kinematics (FK) from $\mathbf{q}_\text{current}$
    \State \textbf{Leader Arm:} Compute deviation $\Delta \mathbf{ee} = \mathbf{ee}_\text{target} - \mathbf{ee}_\text{current}$
    \State \textbf{Leader Arm:} Compute force feedback factor:
        \[
        \text{Factor} = \sqrt{\frac{\alpha \cdot \|\Delta \mathbf{ee}\|^2}{1 + \|\mathbf{v}_\text{cartesian}\|^2}}
        \]
    \State \textbf{Leader Arm:} Update virtual target pose and impedance gains $K_p$, $K_d$ using the force feedback factor
\Until{task complete}
\end{algorithmic}
\end{algorithm}

\section{Experiments}

\subsection{Experiment Design.} 
ACE-F was evaluated according to its performance controlling a Franka Emika Panda robot and two variations of the UFactory XArm in virtual and real-world experiments. Additional demonstrations were performed with a wide variety of robot platforms available in the Mujoco based Robosuite simulator to demonstrate ACE-F's generalizable nature \cite{robosuite2020}. Joint-copy methods, like Gello, were used as a baseline to demonstrate ACE-F's intuitive design for new users, and the advantages that the simplified structure provides. Further testing trained imitation learning models on data collected using ACE-F with and without force-feedback enabled in order to highlight tasks in which force feedback is especially useful.

\subsubsection{Ablation Study} Before performing the simulated and real world experiments, we conducted an ablation study to explore how different substitutions of the velocity term in the feedback compensation formula impact system stability and user experience. Specifically, we tested four variations: absolute value $|v|$, squared velocity $v^2$, exponential $\exp(v)$, and hyperbolic tangent $\tanh(v)$. Table~\ref{tab:stability_metrics} shows that $|v|$ produced the lowest high frequency energy ratio (0.123), lowest maximum local jerk (0.00064), and highest feedback correlation (0.758), indicating it provides the most stable and precise force feedback. In contrast, $\exp(v)$ introduced the highest high frequency energy ratio (0.211) and jerk anomalies (4.59\%), suggesting a more aggressive but potentially destabilizing feedback response. $\tanh(v)$ and $v^2$ offered intermediate results, with $tanh(v)$ achieving a feedback correlation of 0.662 and relatively low jerk anomalies (0.81\%). These findings demonstrate the trade-offs between stability, responsiveness, and user sensitivity across different velocity formulations in the feedback term.

\begin{table*}[ht]
  \centering
  \caption{Teleoperation Stability Metrics for Different Velocity Transformations}
  \label{tab:stability_metrics}
  \begin{tabular}{lcccc}
    \toprule
    Feedback Param & High Freq. Energy Ratio & Max Local Jerk & Leader-only Jerk Anomaly (\%) & Feedback Correlation \\
    \midrule
    $|v|$         & \textbf{0.123} & \textbf{0.00064} & \textbf{0.00} & \textbf{0.758} \\
    $v^2$         & 0.132 & 0.00219 & 0.39 & 0.521 \\
    $\exp(v)$     & 0.211 & 0.00273 & 4.59 & 0.686 \\
    $\tanh(v)$    & 0.131 & 0.00139 & 0.81 & 0.662 \\
    \bottomrule
  \end{tabular}
\end{table*}

\subsubsection{Baseline Simulation Experiments} First, we compared ACE-F and Gello on three contact-rich simulations built in MuJoCo -- Box Stacking, Box Dragging, and Table Mopping. Each task involved varying levels of physical interaction, visual reasoning, and force modulation, which we could use to evaluate the two platforms. Two groups of users with varying levels of teleoperation experience were selected. The groups began each task using a different system and switched platforms halfway through the task. Additionally, users were allowed up to five minutes of practice every time they switched platform. Practice sessions were conducted in a task neutral practice arena, which contained elements from each of the three actual experiments. Users were provided with a description of their goal before each task and loaded into the practice scene. Once they reached the end of their practice time or self-determined that they were ready to begin, they were given control of a Franka Emika Panda robotic arm within the scene and allowed to use any combination of force feedback, visual cues, and tools to complete the task. The camera position and robot remained the same through all three tasks.

\paragraph{Simulated Box Stacking} Four blocks with randomized weights and a balance scale were placed on the table, and users were asked to determine the relative weight of each block before stacking them from heaviest to lightest. Users were graded on four metrics: speed, number of scale uses, number of times they knocked one or more blocks off their tower, and whether or not they were successful in stacking all four blocks in the correct order. 

\paragraph{Simulated Box Dragging} A singular weighted block attached to a handle was placed at one end of the table and an invisible obstacle was randomly placed somewhere on the other side. Users were asked to use the handle to drag the block across the table until they reached the obstacle, at which point their goal was to release it without crossing over the line. They were allowed to use any combination of force feedback and visual cues available in the simulation and were graded on three metrics: speed, distance from the obstacle, and whether or not they were successful in stopping the box before it crossed the line. 

\paragraph{Simulated Table Mopping} Users were tasked with dragging the robot's end-effector back and forth along a line on a table from one edge to the other, while trying to maintain steady speed and pressure. This task was performed on two table configurations -- one where the line went from left to right, and one where the line went from close to far. Users were graded on their speed and force consistency, defined as the variation in normal forces experienced by the table in the simulation.

\subsubsection{Baseline Real-World Experiments} Next, ACE-F and Gello were evaluated on the following real-world tasks: Can Stacking, Marker Erasing, and Hidden Insertion. Users were once again allowed up to five minutes of practice every time they switched platform, and practice sessions were conducted in the same environment where the tests took place.

\paragraph{Real Can Stacking} Similar to the simulated stacking scene, this experiment consisted of three randomly placed weighted cans and a digital balance scale. Users were instructed to determine the relative weight of each can before stacking them from heaviest to lightest. Users were graded on four metrics: speed, number of scale uses, number of times they knocked one or more cans off their tower, and whether or not they were successful in stacking all three cans in the correct order. 

\paragraph{Real Marker Erasing} Similar to the simulated mopping scene, this experiment consisted of a raised platform holding a textured ceramic dish with a 2-inch by 2-inch square of dry-erase marker. Users were instructed to use a whiteboard eraser to remove the visible markings from the plate. Users were graded on three metrics: speed, the number of times they triggered the robot's built-in safety features, and whether or not they were successful in removing all of the marker from the plate. 

\begin{figure}[t]
  \centering
  \includegraphics[width=\columnwidth]{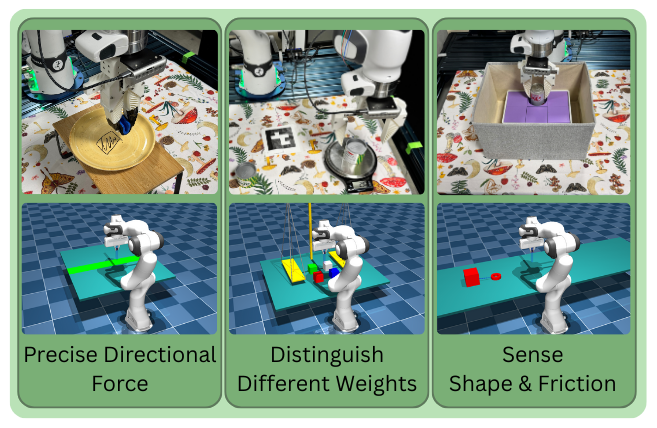}
  \caption{
    Overview of the six tasks in the evaluation suite: real-world mopping, real-world can stacking, real-world blind can insertion, simulated table mopping (left-right and forward-backward), simulated box stacking, and simulated box dragging.
  }
  \label{fig:Gello_comp}
\end{figure}

\begin{table}[t]
\centering
\caption{Aggregate simulation performance across three teleoperation tasks. ACE-F demonstrates advantages over Gello in every task.}
\label{tab:simulated-all-tasks-summary}
\begin{tabular}{@{}lccc@{}}
\toprule
\multicolumn{4}{c}{\textbf{Virtual Stacking Task}} \\
\midrule
\textbf{Method} & \textbf{Avg. Time (s)}& \textbf{Avg. Scale Uses}& \textbf{Success Rate (\%)}\\
ACE-F& \textbf{102.1 ± 27.2}& \textbf{0.7 ± 0.5}& \textbf{90.0}\\
Gello & 187.0 ± 121.2 & 2.3 ± 0.7 & 70.0 \\
\midrule
\multicolumn{4}{c}{\textbf{Virtual Box Dragging Task}} \\
\midrule
\textbf{Method} & \textbf{Avg. Time (s)}& \textbf{Light Status} & \textbf{Success Rate (\%)}\\
ACE-F& 16.7 ± 2.5 & On & \textbf{100}\\
ACE-F& \textbf{16.4 ± 2.5}& Off & 94.4 \\
Gello & 21.9 ± 4.8 & On & 75.0 \\
\midrule
\multicolumn{4}{c}{\textbf{Virtual Mopping Task: Left-Right}} \\
\midrule
\textbf{Method} & \textbf{Normal Force} & \textbf{Max/Avg. Force Ratio}& \\
ACE-F& \textbf{196.8 ± 55.9}& \textbf{4.59}& \\
Gello & 231.9 ± 69.6& 7.19& \\
\midrule
\multicolumn{4}{c}{\textbf{Virtual Mopping Task: Forward-Backward}} \\
\midrule
\textbf{Method} & \textbf{Normal Force} & \textbf{Max/Avg. Force Ratio}& \\
ACE-F& \textbf{229.0 ± 146.8}& \textbf{7.19}& \\
Gello & 251.7 ± 183.1& 9.73& \\
\bottomrule
\end{tabular}
\end{table}

\paragraph{Real Blind Can Insertion} This task was performed exclusively in the real-world using a rectangular box with a randomized hole position. It was placed behind a cardboard screen to obscure part of the robot operator's vision of the task space. Users were instructed to pick up a can from the visible portion of the table and place it into a hole located behind the cardboard screen. Users were graded on their speed, the number of times they triggered Franka's built-in safety features, and whether or not they successfully placed the can into the hole.

\subsubsection{Simulated Model Comparisons} The rest of the simulations were conducted in the Robosuite simulator, using their lifting, stacking, and wiping benchmark task environments \cite{robosuite2020}. Lifting consisted of grabbing and lifting a randomly positioned and oriented block, stacking required the policy to place a small block onto a larger block, and wiping required the policy to use the end-effector to remove a streak from a table surface. Each model was trained using two camera views and qpos data obtained from Robosuite over the course of 50 demonstrations. The lifting and wiping tasks were performed with a Franka Emika robot, and the stacking task used a UFactory Xarm7. Each task policy was trained once with force feedback enabled and once without it in order to evaluate its importance. The results of the first 20 task attempts were recorded while the demonstrations were collected as an additional performance metric. Importantly, data was collected without an intermediate motion planner, explaining the slightly below average task performance.

\subsubsection{Real-World Model Comparisons} Real-world policy deployments were performed on a Franka Emika Panda robot. Similar to the simulated examples, the policy was trained with two camera views and qpos data collected by a human operator with force feedback enabled. Franka was trained to sort a random series of cans into three slots in increasing size. It was trained on 18 episodes containing three examples of each starting combination of cans.

\subsection{Experimental Results.} 

\subsubsection{Baseline Simulation Experiments} 

\paragraph{Simulated Box Stacking} ACE-F demonstrated a clear advantage over Gello in speed, number of scale uses, number of tower topples (labeled as blunders in Table~\ref{tab:simulated-all-tasks-summary}), and success rate. ACE-F testers able to complete the stacking task 54.62\% faster than Gello users with far more consistency (a quarter the standard deviation). Additionally, ACE-F users were 28.57\% more successful, despite using the scale less than half as much as Gello users. Since ACE-F allows users to feel the weight of the cube without having to rely on the scale, users spent less time weighing each block, so they could limit their motions to a smaller area of the task space. The reduced motion and ACE-F's additional sense of touch explains why ACE-F users made fewer blunders during stacking.

\paragraph{Simulated Box Dragging} ACE-F performed significantly better than Gello in the box dragging task, where it was 23.72\% faster, even without vision for the latter half of the task. It was also far more successful, only failing the task in 5.6\% of the blind tasks, compared with Gello's 25\% failure rate with the lights on. This test highlights ACE-F's advantages in tasks where tactile feedback can make up for visual cues.

\begin{table}[t]
\centering
\caption{Aggregate real-world performance across three teleoperation tasks. ACE-F consistently outperforms the joint-copy Gello method in success rate and stability, while reducing reliance on external tools like scales.}
\label{tab:realworld_alltasks_summary}
\begin{tabular}{@{}lccc@{}}
\toprule
\multicolumn{4}{c}{\textbf{Real Stacking Task}} \\
\midrule
\textbf{Method} & \textbf{Avg. Time (s)} & \textbf{Avg. Scale Uses} & \textbf{Success Rate (\%)}\\
ACE   & 90.81 ± \textbf{17.97}& \textbf{0.0 ± 0.00}& \textbf{83.3}\\
Gello & \textbf{88.36} ± 23.46& 2.0 ± 0.63 & 66.7 \\
\midrule
\multicolumn{4}{c}{\textbf{Real Erasing Task}} \\
\midrule
\textbf{Method} & \textbf{Avg. Time (s)} & \textbf{\# Safety Warnings} & \textbf{Success Rate (\%)}\\
ACE   & 26.54 ± 7.89 & \textbf{0}& 100.0 \\
Gello & \textbf{22.13 ± 4.30}& 1 & 100.0 \\
\midrule
\multicolumn{4}{c}{\textbf{Real Blind Can Insertion Task}} \\
\midrule
\textbf{Method} & \textbf{Avg. Time (s)} & \textbf{Success Rate (\%)}& \\
ACE   & 43.26 ± 20.87 & \textbf{100.0}& \\
Gello & \textbf{34.42 ± 12.30}& 50.0 & \\
\bottomrule
\end{tabular}
\end{table}

\paragraph{Simulated Table Mopping} ACE-F outperformed Gello in both configurations of the mopping task, as well. By providing the user with an even force when the arm collides with the table, the users could sense how much force they were applying and more easily regulate their downward pressure. This is clearly indicated by a 36.16\% smaller maximum force to average force ratio in the left-right configuration and a 41.86\% smaller ratio in the forward-backward test.

\subsubsection{Baseline Real-World Experiments} 

\paragraph{Real Can Stacking} Gello performed the can stacking task faster on average, however it had a larger standard deviation in completion times and a lower success rate. This can be attributed to two things: the platform's unstable configuration and the increased movements associated with moving cans to the scale. ACE-F users did not have to use the scale in any of their trials because they could feel the can's weight purely through force feedback. Thus, they could avoid making large movements over to the scale, which was more likely shake the can loose from its gripper. Similarly, Gello's structure makes it difficult for the user to keep the gripper perfectly upright, which contributed to the cans falling from its grasp more frequently. If the can landed in an awkward orientation, it could roll outside of the robot's workspace, causing it to immediately fail.

\paragraph{Real Marker Erasing} Both ACE-F and Gello had a 100\% success rate in this task, however Gello performed 16.62\% faster than ACE-F on average and had a smaller standard deviation in its times. This is partially due to the fact that Gello is joint-copy, and thus users can quickly move multiple joints at their maximum speeds simultaneously. This accelerated their performance but also triggered a warning in Franka's safety system during one of the tasks.

\paragraph{Real Blind Can Insertion} ACE-F performed significantly better than Gello in this task because the force feedback allowed users to compensate for their poor vision by feeling around inside the box. Gello users generally completed this task faster, however they possessed a much lower success rate of 50\%, while ACE-F users correctly identified the can's target location 100\% of the time.

\begin{table}[t]
\centering
\caption{Imitation learning policy performance across three teleoperation tasks. ACE-F with force feedback (FF) demonstrates advantages over ACE-F without FF in every task.}
\label{tab:imitation-all-tasks-summary}
\begin{tabular}{@{}l>{\raggedright\arraybackslash}p{0.2\linewidth}>{\centering\arraybackslash}p{0.2\linewidth}>{\centering\arraybackslash}p{0.25\linewidth}@{}}
\toprule
\multicolumn{4}{c}{\textbf{Data Collection}}\\
\midrule
\textbf{Task}&  \textbf{Environment}&\textbf{Success Rate (\%)}& \textbf{Avg. Time (s)} \\
FF Lift&  Simulation&96.0&  \textbf{7.0 ± 0.7}\\
 Regular Lift&  Simulation&96.0& 8.9 ± 1.3\\
\cmidrule(lr){1-4}   
 FF Stack&  Simulation&\textbf{95.8}& \textbf{9.5 ± 1.0}\\
 Regular Stack&  Simulation&72.0& 9.9 ± 1.5\\
\cmidrule(lr){1-4}   
 FF Wipe&  Simulation&\textbf{100.0}& \textbf{6.9 ± 1.8}\\
Regular Wipe&  Simulation&76.0&  10.1 ± 3.1\\
\midrule
\multicolumn{4}{c}{\textbf{Policy Evaluation}}\\
\midrule
\textbf{Task}&  \textbf{Environment}&\textbf{Full Success Rate (\%)}& \textbf{Partial Success Rate (\%)}\\
 FF Lift&  Simulation&\textbf{95.0}&N/A\\
 Regular Lift
&  Simulation&60.0&N/A\\
\cmidrule(lr){1-4}   
FF Stack&  Simulation&\textbf{52.6}&  \textbf{44.4}\\
 Regular Stack
&  Simulation&30.0& 21.4\\
\cmidrule(lr){1-4}   
FF Wipe&  Simulation&\textbf{75.0}&  N/A\\
 Regular Wipe&  Simulation&65.0& N/A\\
\cmidrule(lr){1-4}   
 FF Can Sort&  Real-World&66.7& 75.0\\
\end{tabular}
\end{table}

\subsubsection{Simulated Model Experiments} 

\paragraph{Simulated Lifting Policy} ACE-F performed comparably with and without force feedback during data collection, however users were slightly faster with force feedback enabled, likely because they could use the collision with the table as an indicator of when to close the gripper. Surprisingly, the force-enabled policy performed significantly better than the force-disabled policy. One possible reason is that the actions recorded in the force-enabled dataset were far more consistent than those in other training data, leading to a higher policy confidence.

\paragraph{Simulated Stacking Policy} Force feedback provided a large advantage in the stacking task demonstrations because it served as a depth indicator to users in both the pick and place portions of the task. This corresponds with higher success rates in the trained policy, though the overall values are slightly below average without an intermediate motion planner. In this case, full successes required successful completion of both pick and place actions, and 
\begin{figure}[H]
  \centering
  \includegraphics[width=\columnwidth]{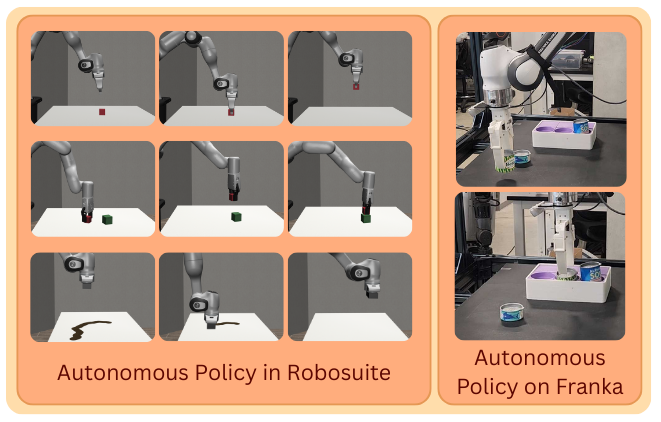}
  \caption{
    Overview of all four imitation learning policies: simulated lifting, simulated stacking, simulated wiping, and real-world can sorting.
  }
  \label{fig:policy_comp}
\end{figure}
partial successes were counted when the robot successfully picked up the block but could not place it.

\paragraph{Simulated Wiping Policy} The wiping task showed the largest improvements when completed with force feedback because users were able to regulate how much force they applied to the table. This prevented users from pressing into the table so hard that the second to last arm joint buckled and triggered a joint limit error, which was the primary failure method in this task. The force-enabled policy also performed better than its peer but to a smaller degree. Notably, both policies exhibited the highest failure rates near the edges of the randomized task space, likely due to out-of-distribution states.

\subsubsection{Real-World Model Experiments} 

\paragraph{Real Sorting Policy} This task successfully trained a policy in the real-world, and although it does not have a baseline comparison without force feedback, the data in previous sections implies that the advantages of force feedback would carry over. Stopping short of making this claim, however, this test demonstrates that ACE-F can collect high-quality data in the real-world.

\section{Related Work}
\subsection{Force Feedback Teleoperation.} Force feedback has become a widely recognized enabler for contact-rich teleoperation, allowing operators to perceive interaction forces and improve manipulation performance~\cite{Miller2021impact,ElRassi2020review, Ding2024bunny}. 
Although many commercial robot arms incorporate 6‐DoF FT sensors, their high cost and integration complexity make them impractical for general deployment~\cite{Ubeda2018design}. Furthermore, the majority of low‐cost teleoperation systems forgo force feedback entirely, relying solely on position or joint commands with no haptic cues~\cite{Zhao2023learning, wu2023gello, Fu2024mobile, yang2024ace}. Traditional force‐feedback teleoperation implementations therefore depend on external FT sensors mounted on the follower device, imposing hardware and calibration burdens~\cite{Ehrampoosh2022force, bazhenov2025echo, Liu2025factr}. 
Virtual force feedback schemes approximate contact forces via kinematic or impedance models, but cannot fully capture true interaction dynamics~\cite{Abiri2019multi}. Some teleoperation systems have also attempted to convey force information through non-haptic channels—e.g., visual overlays on the video stream, audio alerts, or controller vibration cues—but these indirect modalities often lack intuitiveness and can increase operator cognitive load~\cite{ElRassi2020review}.
To overcome these limitations, ACE-F infers real-time 3-DoF end-effector forces from trajectory deviations, providing sufficiently accurate force cues for daily-life teleoperation tasks without any additional sensors.

\subsection{Cross-Embodiment Teleoperation.} Mapping human motions to robots with differing kinematic structures is necessary for proper teleoperation~\cite{lv2022gulim}. Direct joint-copying approaches build a small leader arm or mobile controller that mirrors the target robot’s kinematics, providing intuitive, low-latency mapping, but require rebuilding the hardware for each new robot~\cite{Zhao2023learning, wu2023gello, Fu2024mobile, bazhenov2025echo, Liu2025factr,Ze2025TWIST}. In contrast, IK-based Cartesian control naturally generalizes across embodiments, allowing a single leader interface to drive robots of varied morphologies without any hardware changes~\cite{Qin2023AnyTeleop, zhao2023wearable, yang2024ace, kumar2015mujoco,qin2022one,handa2020dexpilot,aronson2022gaze, kofman2005teleoperation}. IK-driven teleoperation systems typically use four main interfaces to obtain wrist and hand pose: motion-capture devices~\cite{wang2024dexcap, fan2023arctic,taheri2020grab, liu2019high,caeiro2021systematic}, cameras~\cite{Qin2023AnyTeleop}, VR equipment\cite{han2020megatrack,arunachalam2023dexterous,kumar2015mujoco, Ebert2021bridge, Cheng2024opentelevision, Li2025AMO, Lu2024mobile, jiang2025gsworld}, or exoskeleton hardware~\cite{zhao2023wearable, yang2024ace, Ben2025homie, Fang2023AirExo, toedtheide2023force,ishiguro2020bilateral, lee2014robot, buongiorno2019multi}. The first three approaches can capture complete wrist and hand information to enable dexterous end-effector control~\cite{zhang2020mediapipe, ohkawa2023assemblyhands, handa2020dexpilot, Qin2023AnyTeleop,li2020mobile}, but their interfaces make integrating force feedback difficult. Exoskeleton-based systems offer a direct way to add force feedback, yet their bulky size and mechanical complexity significantly increase torque requirements—raising motor costs and reducing wearability. By contrast, ACE-F combines a compact, foldable 3-DoF leader arm with glove-based hand tracking to achieve precise, occlusion-free full-hand pose capture while minimizing device volume and torque demands—thereby lowering motor performance requirements and overall system cost, and enabling seamless integration of tactile gloves in the future. 

\section{Discussion}
\label{sec:conclusion}
ACE-F displayed clear advantages over joint-copy methods, like Gello, when compared in virtual and real-world environments. It performed better in most tasks because its inverse-kinematic controller removes the burden of monitoring the robot's configuration from the operator and the force-feedback from the inverse-dynamics controller gives the user an extra sense, which improved environmental awareness. Additionally, by reducing the complexity of the overall system, ACE-F remains very compact and portable. The sensor-less force feedback also enables users to complete tasks in new domains, where the user has limited vision of their workspace. This solves one of the major drawbacks of joint-copy methods and results in a surprise benefit by limiting user speed through resistance forces, which reduces the likelihood of the user triggering speed-based warnings. Lastly, imitation learning models trained on data from ACE-F demonstrated clear improvements from the implementation of force feedback in simulation, and the same pipeline can be applied to data collection for real-world policies. ACE-F comes with its own drawbacks, however. For instance, better motors could supply stronger feedback to the user, and this project prioritizes low-cost teleoperation over full 6-DoF force feedback, since cartesian forces are generally considered sufficient for most tasks. That leaves a gap in the scope of the project, which can be improved in future releases. Future works should focus on expanding the implementation of the gloves and incorporating torque feedback for the user in addition to the cartesian forces currently provided.

\bibliographystyle{IEEEtran}
\bibliography{main}

\end{document}